\documentclass[sn-mathphys-num, iicol, pdflatex]{sn-jnl}% Math and Physical Sciences Numbered Reference Style 
\usepackage{graphicx}%
\usepackage{multirow}%
\usepackage{amsmath,amssymb,amsfonts}%
\usepackage{amsthm}%
\usepackage{mathrsfs}%
\usepackage[title]{appendix}%
\usepackage{xcolor}%
\usepackage{textcomp}%
\usepackage{manyfoot}%
\usepackage{booktabs}%
\usepackage{algorithm}%
\usepackage{algorithmicx}%
\usepackage{algpseudocode}%
\usepackage{listings}%
\usepackage{subcaption} % To facilitate creation of subfigures
\usepackage{verbatim} % block comments
\usepackage{dsfont} % indicator function

\newtheorem{definition}{Definition}%

\begin{document}

\title[Article Title]{Generating the Ground Truth: Synthetic Data for Soft Label and Label Noise Research}


\author*[1,2]{\fnm{Sjoerd} \spfx{de} \sur{Vries}}\email{s.devries1@uu.nl}

\author[1]{\fnm{Dirk} \sur{Thierens}}

\affil*[1]{\orgdiv{Department of Information and Computing Sciences}, \orgname{ Utrecht University}, \orgaddress{ \street{Princetonplein 5}, \postcode{3584 CC}, \city{Utrecht}, \country{The Netherlands}}} 

\affil[2]{\orgdiv{Department of Digital Health}, \orgname{University Medical Center Utrecht}, \orgaddress{ \street{Heidelberglaan 100}, \postcode{3584 CX}, \city{Utrecht}, \country{The Netherlands}}}  

\abstract{In many real-world classification tasks, label noise is an unavoidable issue that adversely affects the generalization error of machine learning models. Additionally, evaluating how methods handle such noise is complicated, as the effect label noise has on their performance cannot be accurately quantified without clean labels. Existing research on label noise typically relies on either noisy or oversimplified simulated data as a baseline, into which additional noise with known properties is injected. In this paper, we introduce SYNLABEL, a framework designed to address these limitations by creating noiseless datasets informed by real-world data. SYNLABEL supports defining a pre-specified or learned function as the ground truth function, which can then be used for generating new clean labels. Furthermore, by repeatedly resampling values for selected features within the domain of the function, evaluating the function and aggregating the resulting labels, each data point can be assigned a soft label or label distribution. These distributions capture the inherent uncertainty present in many real-world datasets and enable the direct injection and quantification of label noise. The generated datasets serve as a clean baseline of adjustable complexity, into which various types of noise can be introduced. Additionally, they facilitate research into soft label learning and related applications. We demonstrate the application of SYNLABEL, showcasing its ability to precisely quantify label noise and its improvement over existing methodologies.}


\maketitle

\section{Introduction}\label{sec:introduction}

Classification models are of great interest to both the research community and machine learning (ML) practitioners. When applied to real-world problems, these models encounter noisy data, where noise is defined as anything that obscures the relationship between the dependent and independent variables~\cite{hickey1996noise}. Label noise, in particular, can adversely impact classifier performance, model complexity, learning rates and effect size estimation~\cite{frenay2014classification}. Consequently, a lot of research is focused on developing prediction methods that are robust to label noise and preprocessing steps for filtering such noise from data~\cite{frenay2014classification,song2022learning}.
 
Typically, when assessing the robustness of an algorithm to label noise, researchers designate a dataset as the ground truth and introduce artificial noise into the labels. This dataset can be an existing real-world dataset, either unaltered or with label noise filtering techniques applied to it~\cite{berthon2021confidence,cheng2020learning}. Alternatively, data is entirely simulated, often by modelling relatively simple relationships between the dependent and independent variables~\cite{berthon2021confidence,cheng2020learning}. Lastly, a limited number of curated datasets are publicly available for which the label noise has been quantified by expert annotators~\cite{xiao2015learning, lee2018cleannet, li2017webvision}.

Each of these approaches has inherent drawbacks: For real-world datasets there are no clean labels available for evaluation, requiring noise to be injected into already noisy data. Alternatively, attempts are made to clean the label noise from the data using filter methods. These either reduce the size of the dataset by removing uncertain instances or try to correct the existing labels, a process that is inherently uncertain. The relationships in simulated datasets are often overly simplistic, making them unrealistic representations of real-world data. Finally, for curated data, the noise cannot be tailored, limiting the performance evaluation of a method to one very specific noise pattern. Creating a curated dataset takes considerable effort as well. While noisy real-world data are abundant, the absence of clean evaluation sets remains a significant challenge. Fr\'enay and Verleysen state that, when comparing methods, the presence of label noise in the validation data causes any estimates to be off by an unknown amount~\cite{frenay2014classification}. They highlight this as an important open research question, which we address in this work.

In this paper, we aim to advance existing experimental strategies by introducing the SYNLABEL framework: Synthetic Labels As Baseline for Experiments with Label Noise. SYNLABEL facilitates the construction of artificial tabular ground truth datasets that may be used as baselines for evaluating method performance in the presence of label noise. Additionally, SYNLABEL incorporates the Feature Hiding method, which transforms a ground truth dataset with hard labels into a dataset with soft labels. A hard label is definitive assignment of an instance to a single class, commonly used in ML datasets. In contrast, a soft label is a probability vector over the possible classes, indicating the likelihood that an instance belongs to each class, thereby allowing for the expression of uncertainty.

To generate a hard labelled ground truth dataset, SYNLABEL defines a prediction model trained on any dataset as the ground truth function. By applying this ground truth function to the original data or any other data within its domain, associated noiseless labels are generated. As our focus is on testing methods under known noise conditions, rather than finding the best model for a specific real-world problem, this function need not perfectly represent the original data. The generated ground truth set can be further transformed into a partial ground truth set for which each data point is accompanied by a soft label using Feature Hiding. This method involves hiding a number of specified variables from the ground truth dataset. Subsequently, by learning or specifying a (conditional) prior distribution for the hidden variables, resampling values from it and combining these multiple sampled values with the fixed values for the variables that were not hidden, a posterior distribution is generated via the ground truth function. Although the prior distribution over the hidden variables is almost guaranteed not to match the exact underlying distribution of the real-world generative distribution over the original dataset, this is unnecessary for our purpose.

The advantage of using a dataset with soft labels as a baseline, as opposed to one with hard labels, is that it allows for explicit quantification and direct injection of label noise. Furthermore, these datasets can be used to simulate problems from different domains that can be mapped to a label distribution problem, such as learning from crowds or with confidence scores. The constructed sets with hard or soft labels, that do not contain any noise, serve as a starting point to which further transformations can be applied in order to generate the specific noise of interest.

In summary, the key contributions of this paper are:

\begin{itemize}
\item The SYNLABEL framework, which facilitates the generation of experimental datasets for soft label and label noise research.
\item A method for constructing a ground truth dataset informed by real-world data for evaluation purposes.
\item A method for converting hard into soft labels by resampling values for features hidden from the ground truth function.
\item Analyses showing the benefits of using Feature Hiding for introducing uncertainty into hard labels and the advantages of using the resulting soft labels for the quantification and injection of label noise.
\end{itemize}

\section{Related Work}\label{sec:related_work}

Systematic experiments in label noise research call for datasets with both clean labels as a baseline for evaluation, as well as noisy labels. Based on the methods for obtaining these sets, experiments in the field of label noise can be placed into three categories: (1) an existing dataset for which the labels have been manually corrected is used, (2) artificial noise is injected into a clean dataset which has been simulated or (3) noise is injected into an existing real-world dataset.

The first category involves curated real-world datasets for which noisy labels have been corrected, such as Clothing1M~\cite{xiao2015learning}, Food-101N~\cite{lee2018cleannet} and WebVision~\cite{li2017webvision}. The ground truth labels are typically determined by a panel of experts. However, even among experts high inter-observer disagreement may occur~\cite{veta2016mitosis}. 

In the second and third categories, a baseline dataset is first established, after which noise is injected into the labels based on a chosen noise model. The injected noise can be categorized, in order of decreasing commonality and increasing complexity, as Noisy Completely At Random (NCAR), Noisy At Random (NAR) or Noisy Not At Random (NNAR)~\cite{frenay2014classification}. In the NCAR model, the noise does not depend on either the dependent or the independent variables. The labels are corrupted randomly~\cite{angluin1988learning}. The NAR model, or class conditional model, takes the true class of an instance into account when producing a noisy label, by means of a transition matrix~\cite{lawrence2001estimating}. The NNAR model is the most complex, allowing for the label noise to be based on both the dependent and independent variables, making it instance-dependent. An example of this type of label noise is provided by Berthon et al.~\cite{berthon2021confidence}, who simulated a dataset of concentric circles and varied the magnitude of the added noise to a point based on its location in the plane. In the same work, they use real-world data to which they add noise based on the output of a classifier, a common method for introducing label noise via the NNAR model~\cite{jin2002learning,chen2021beyond,xia2020part,zhang2021learning,garcia2019new,gu2022instance}. 

In the second category, when synthetic data is constructed, the true underlying function is known by design, e.g. the data is sampled from Gaussian distributions or constructed using rule-based generation~\cite{hickey1996noise}. Clean labels are generated from this function, to which noise can be added. However, these simulated datasets often lack the complex interactions between variables present in real data. 

In the third category, real-world data, the true labels remain unknown when there are no resources available for curation and thus the level of noise in the baseline cannot be quantified~\cite{hickey1996noise}. The importance of using controllable artificial data, especially in the context of noise, was already mentioned in~\cite{langley1988machine}, as it enables systematic research into different aspects of a domain. While using real-world data is the least labour intensive experimental method, the effect that any noise added to the labels has on method performance cannot be separated from the effect due to the inherent noise present in the data.

In summary, while different types of label noise experiments exists, each has its limitations. Moreover, when using hard labels, label noise specification for an individual label is binary -- either the label is correct or not. 

Recently, Gu et al.~\cite{gu2022instance} proposed a framework for generating instance-dependent label noise. Our work differs from theirs in several ways: while they introduce a specific type of classifier-based instance-dependent noise, our framework supports any type of noise injection. Furthermore, their approach requires a dataset with clean labels as a starting point, whereas our methods can be used for generating such noiseless datasets with either hard or soft labels.

Beyond label noise research, SYNLABEL is also suitable for generating datasets to evaluate soft label methods. Several research fields work with soft labels directly or can be adapted to fit the soft label learning paradigm. The terms label distribution learning~\cite{geng2016label} and multiple-label learning~\cite{jin2002learning} are used interchangeably with soft label learning. An example of a task that can be transformed into a soft label learning task is learning from labels assigned by human annotators, who could be either experts or a crowd~\cite{zhang2016learning, dizaji2020robust}. Multiple hard labels are obtained per instance, which can be aggregated to produce soft labels~\cite{jin2002learning, barsoum2016training}. Alternatively, these annotators can provide a confidence score with their hard label, i.e. a measure of their uncertainty, which can be transformed into a soft label~\cite{nguyen2014learning,oyama2013accurate}. Techniques that try to estimate the best matching soft labels for a dataset with hard labelled instances are called Label Enhancement (LE), and these include include Fuzzy Clustering LE~\cite{el2006study} and Graph Laplacian LE~\cite{xu2019label}.

\section{The SYNLABEL Framework}\label{sec:framework}

\begin{figure}[!t]
    \centering
    \includegraphics[width=0.47\textwidth]{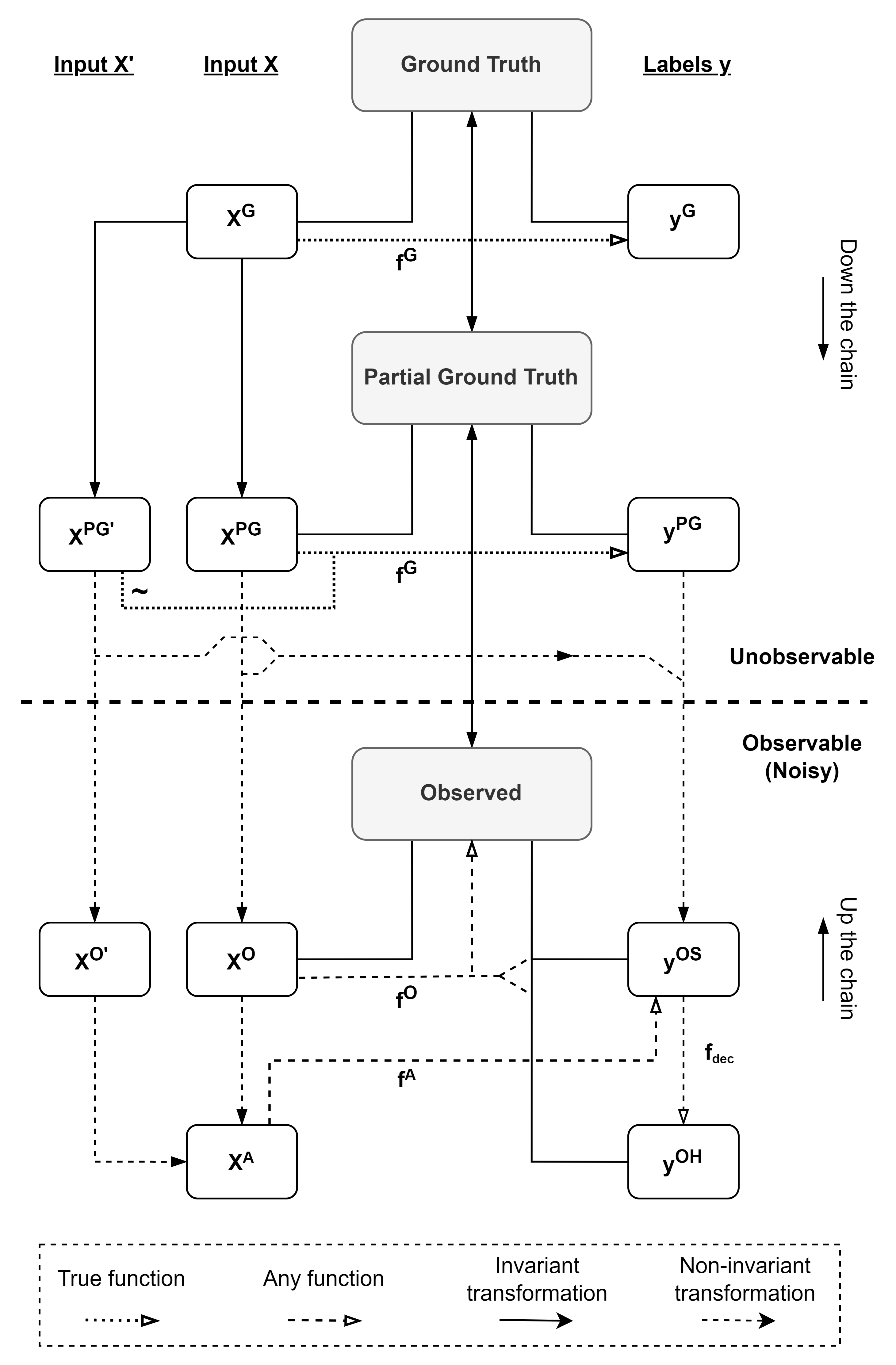}
    \caption{A schematic overview of the SYNLABEL framework. The white boxes represent data, either input $X$ or labels $y$. The gray boxes represent a type of dataset, linked by a solid line to their input and output. The arrows represent the different transformations and functions defined by the framework. $\sim$: sampled. $f$: a function.}
    \label{fig:framework}
\end{figure}

We present the Synthetic Labels As Baseline for Experiments with Label noise (SYNLABEL) framework, depicted in Figure~\ref{fig:framework}, which facilitates generating synthetic tabular datasets for label noise experiments. SYNLABEL defines various dataset types and transformations between them. A dataset, denoted as $D$, is defined as $D = (X,y) = \lbrace z_i \rbrace_{i=1}^n = \lbrace x_i, y_i \rbrace_{i=1}^n$. Here, $(X,y)$ is the notation used on the dataset level, where $X$ is the input matrix with domain $\mathcal{X}$ and $y$ are the labels with domain $\mathcal{Y}$, either a label vector for hard labelled dataset types or a label matrix for soft labelled dataset types (distinguished by their superscript as defined in Definition~\ref{def:G},\ref{def:PG},\ref{def:OS} and~\ref{def:OH}). $z_i$ denotes a single instance $i\in \lbrace 1,2,..,n \rbrace$ in the dataset with domain $\mathcal{Z} = \mathcal{X} \times \mathcal{Y}$, and it consists of the input vector $x_i$ and label $y_i$, which is a scalar for hard labels or a vector for soft labels, depending on the dataset type. 

Additionally, SYNLABEL defines two types of ground truth datasets that include in their definition a deterministic function $f^G: \mathcal{X} \rightarrow \mathcal{Y}$ that establishes a noiseless relationship between the inputs and labels: $y_i = f^G(x_i)$. These two types of ground truth datasets, depicted in the Unobservable part of Figure~\ref{fig:framework}, are generally unobtainable for real-world problems, as that would require learning $f^G$ from the data such that it produces the exact labels for the entire domain $\mathcal{X}$. In practice, an Observable dataset is available for a given classification task and often we can only hope to obtain a function $f^O: \mathcal{X} \rightarrow \mathcal{Y}$ that generalizes reasonably well. The datasets can be further categorized based on whether their labels consist of a single unambiguous class $c \in \mathcal{Y}$ for each instance, known as a hard label, or a discrete probability distribution over $\mathcal{Y}$, referred to as a soft label.

Users are encouraged to utilize the framework to construct a noiseless ground truth dataset with either hard or soft labels based on a pre-specified function, such as a trained model. Any additional noise applied to this dataset can be precisely quantified for each data point. This allows for analysing method performance for a particular type of noise in isolation. We emphasize that SYNLABEL is not intended for optimizing models for a specific dataset, but rather for evaluating and comparing methods in the presence of label noise. In the following we describe the different dataset types incorporated in the framework.

\subsection{Dataset Types}\label{sec:notation}

The SYNLABEL framework is intended to be applied to a deterministic classification task:

\vspace{5px}

\begin{definition}\label{def:1}
A deterministic classification task is a task for which, given that the values for all features that are required to determine the outcome $y_i^{G}$ are contained in $x_i^G$ , there is a deterministic true function $f^{G}: \mathcal{X} \rightarrow \mathcal{Y}$ that uniquely defines $y_i$ as $f^G(x_i)$ for its entire domain.
\end{definition}

\vspace{5px}

\noindent In other words, given that we know all of the relevant information $X$ to a task, $y$ is unambiguously determined through the function $f^{G}$. A dataset with a corresponding classification task for which Definition~\ref{def:1} holds is defined as a Ground Truth ($G$) dataset $D^{G}$:

\vspace{5px}

\begin{definition}\label{def:G}
A Ground Truth ($G$) dataset $D^{G} = (X^G, y^G) = \lbrace z_i^G \rbrace_{i=1}^n = \lbrace x_i^G, f^G(x_i^{G}) = y_i^G \rbrace_{i=1}^n$ is a dataset for which any input $x^G_i$ in the domain of $f^G$ is mapped to its deterministic hard label $y^G_i$ by the true function $f^G$.
\end{definition}

\vspace{5px}

\noindent Note that a dataset rarely satisfies this definition unless it is simulated. SYNLABEL offers the tools to generate such a dataset based on noisy, real-world data.

When some of the features required for deterministic classification are unavailable, yet the true classification function is known, a Partial Ground Truth ($PG$) dataset $D^{PG}$ can be obtained, defined as:

\vspace{5px}

\begin{definition}\label{def:PG}
A Partial Ground Truth ($PG$) dataset $D^{PG} = (X^{PG}, y^{PG}) = \lbrace z_i^{PG} \rbrace_{i=1}^n = \lbrace x_i^{PG}, FH(f^G, x_i^{PG}, p_{X^{PG'}}) = y_i^{PG} \rbrace_{i=1}^n$ is a dataset for which any input $x^{PG}_i$ in the domain of $f^G$ is mapped to its soft label $y^{PG}_i$ by the true function $f^G$ through the Feature Hiding ($FH$) method using $p_{X^{PG'}}$, the probability density function of $X^{PG'}$.
\end{definition}

\vspace{5px}

\noindent The intuition behind $D^{PG}$ is that when there are any unavailable (hidden) features, contained in $X^{PG'} \subset X^{G}$, the Ground Truth function $f^G$ misses the information contained in these features, causing uncertainty in the labels $y^{PG}$. Here we employ the subset notation $X^{PG} \subseteq X^{G}$ to denote that $X^{PG}$ contains the columns of $X^G$ corresponding to the subset of variables that is available, while any unavailable (i.e. hidden) features are contained in $X^{PG'} \subset X^{G}$ such that the union of these two equals the original input matrix: $X^{PG} \cup X^{PG'} = X^{G}$. To simulate the uncertainty in the labels, the Feature Hiding ($FH$) method is used to transform a $D^G$ into $D^{PG}$ and generate the corresponding soft labels. As $FH$ is a transformation between datasets, it is further explained in Section~\ref{sec:FH} and the formula for obtaining the probability for each class contained in the soft label is defined in Equation~\ref{eq:soft_label}.  

\begin{table}[!t]
\small
\centering
    \begin{tabular}{lrr} 
        \hline
        \\ [-8pt]
        Dataset type  & Label & Function \\
        \\ [-9pt]
		\hline
		\\[-8pt]
        Ground Truth ($G$) 			& Hard & True \\
        Partial Ground Truth ($PG$) 	& Soft & True \\
        Observed Soft Label ($OS$) 	& Soft & Any \\
        Observed Hard Label ($OH$)			& Hard & Any\\
        \\ [-9pt]
		\hline
    \end{tabular}
\caption{The different dataset types defined in SYNLABEL.}
\label{tab:dataset_restrictions}
\end{table}

The datasets $D^{G}$ and $D^{PG}$ are typically unobtainable in practice as they require that the input and labels are determined exactly by $f^G$, i.e. they may not contain any noise, which is impossible to verify if $f^G$, which can only be estimated, is not known. Data observed in practice, $D^O$, has different characteristics: noise can be present in both the observed input data $X^O$ and the corresponding labels $y^O$. Often, $y^O$ is not measured directly and is instead labelled by an annotator such as an expert, crowd or automated system, based on their own non-deterministic, noisy functional relationship $f^A$. An annotator may have the same information available to them as is available for the classification task, i.e. $X^{A} = X^O$, or additional relevant information $X^{O'}$ could be available: $X^{A} \subseteq (X^O \cup X^{O'})$.

An annotator either implicitly (as an intermediate step in their decision process) or explicitly assigns probabilities to the different classes in $\mathcal{Y}$ for an instance $z_i$, producing a Observed Soft Label (OS) set with corresponding outcome $y_i^{OS}$. This set is often discretized into a Observed Hard Label (OH) set with label $y_i^{OH}$ based on some decision function $f_{dec}$. Depending on whether the intermediate label distributions are preserved ($y^{OS}$) or not ($y^{OH}$), the final dataset becomes either $D^{OS}$:

\vspace{5px}

\begin{definition}\label{def:OS}
An Observed Soft Label ($OS$) dataset $D^{OS} = (X^{O}, y^{OS}) = \lbrace z_i^{OS} \rbrace_{i=1}^n = \lbrace x^{O}_i, y^{OS}_i\rbrace_{i=1}^n$ is a dataset for which the input $X^O$ is associated with soft labels $y^{OS}$.
\end{definition}

\vspace{5px}

\noindent or more commonly $D^{OH}$:

\vspace{5px}

\begin{definition}\label{def:OH}
An Observed Hard Label ($OH$) dataset $D^{OH} = (X^{O}, y^{OH}) = \lbrace z_i^{OH} \rbrace_{i=1}^n = \lbrace x^{O}_i, y^{OH}_i\rbrace_{i=1}^n$ is a dataset for which the input $X^O$ is associated with hard labels $y^{OH}$.
\end{definition}

\vspace{5px}

\noindent Note that there are no function-related requirements for these sets. While a function $f^O$ may be learned from these data, such a function is nearly guaranteed not to match the true functional relationship for the corresponding deterministic classification task. Furthermore, unless the model was overfitted, $f^O(x^O)$ will typically not return the exact labels that it was fit on. However, such a function can be used to construct a new Ground Truth Set, as explained in Section~\ref{sec:up_the_chain}. An overview of the different datasets defined in Definition 2-5 is presented in Table~\ref{tab:dataset_restrictions}.

\section{Data Transformations}\label{sec:data_transformations}

At the core of the SYNLABEL framework are the different operations enabling the transformation of datasets from one type to another. These operations allow users to obtain both a (Partial) Ground Truth dataset for validation and realistic datasets that contain the specific type of label noise required for an experiment.

Datasets can be transformed in two directions within the framework. The direction in which data is most often transformed, to obtain observed datasets containing varying label noise, is referred to as \textit{down the chain}: Ground Truth $(G) \rightarrow$ Partial Ground Truth $(PG) \rightarrow$ Observed Soft Label $(OS) \rightarrow$ Observed Hard Label $(OH)$. Transformations in this direction ensure that instances across different sets remain \textit{coupled}, meaning that two instances $z_i$ in different sets following a transformation down the chain still represent the same entity and maintain their connection to the ground truth function that governs the ground truth data from which they originate. Transformations in the reverse direction, \textit{up the chain}, are typically used for constructing a new ground truth dataset. In this case, however, a distinct dataset is generated for which the instances are different from those in the original set, as a new ground truth relationship is defined, resulting in decoupled instances.

Instances remain coupled following an arbitrary number of transformations both up and down the chain only when all resulting sets are identical, thus $D^G = D^{PG} = D^{OS} = D^{OH}$. This condition implies that the soft label for an instance, $y^{PG}_i$ and $y^{OS}_i$, have probability 1 for the class in $y^{OH}_i$ and $y^G_i$ and probability 0 for the other classes and that the true functional relationship $f^G$ required for the (Partial) Ground Truth is known. The transformation that fulfils this condition is denoted the \textit{identity transformation}.

Figure~\ref{fig:framework} illustrates the transformations and functions along with their corresponding types. A transformation can either be an \textit{invariant transformation}, through which the data are not altered, or a \textit{non-invariant} transformation which introduces uncertainty or noise to the data. A function can either be a \textit{true function} or some other function, such as a learned function or specified decision function, denoted by \textit{any function}.  

\subsection{Down the chain}
\label{sec:down_the_chain}

The following subsections detail the supported transformations down the chain. These transitions ensure the instances $z$ remain coupled between the different datasets.

\subsubsection{From Ground Truth to Partial Ground Truth}\label{sec:FH}

An invariant transformation connects $X^G$ in $D^G$ with $X^{PG'}$ and $X^{PG}$ in $D^{PG}$. Although the variables contained in $X^G$ retain identical values in $D^{PG}$, they may be split into two subsets, $X^{PG}$ and $X^{PG'}$. The function $f^G$ describes the true relationship between $X$ and $y$ for both types of dataset: $f^G(x^G_i)$ yields the true class of instances $z_i$ with absolute certainty, while $f^{G}(x^{PG}_i)$ equals the true soft label of $z_i$. Crucially, when some information contained in $x^{G}_i$ is contained in $x^{PG'}_i$ and consequently missing from $x^{PG}_i$, the function evaluation becomes ambiguous, a scenario we address shortly.

First we show that to obtain a Partial Ground Truth set with soft labels, $D^{PG}$, $X^{PG'}$ cannot simply be ignored and a function $f^{PG}$ learned on $\lbrace X^{PG}, y^{G} \rbrace$. This follows as unless the variables in $X^{PG'}$ can be excluded from the ground truth dataset as they are exactly determined by the variables in $X^{PG}$ or they are irrelevant to the task -- such that they are not in any way considered by $f^G$ -- the missing information contained in $X^{PG'}$ will cause any function $f^{PG}$ learned from only $X^{PG}$ to be different from $f^{G}$. Consequently, some labels produced by $f^{PG}$ are guaranteed to differ from $y^G$ for some instances in the domain $\mathcal{X}$ of $f^G$. Two such labels for the same instance must both be the true label, which is contradictory, causing the sets to become decoupled. The exception is when $X^{PG'}$ is empty, resulting in the special case $X^{PG} = X^{G}$ and $D^G = D^{PG}$, the identity transformation. In that case, since $f^G$ is known from $D^G$, no function has to be learned.  

In addition to the identity transformation, $D^{PG} = D^{G}$, SYNLABEL includes a transformation called Feature Hiding ($FH$), that maintains the coupling between the instances in $D^G$ and $D^{PG}$ when $X^{PG} \neq X^{G}$ and preserves the ground truth relationship $f^G$. Instead of attempting to learn a function $f^{PG}$ from the known features $X^{PG}$, $f^G$ is used along with a specified distribution over the selected hidden features in $X^{PG'}$ to construct $y^{PG}$ as follows: first a number of values $j$ is resampled for the features contained in $X^{PG'}$ for the input $x_i$ of each instance $z_i$ in accordance with a probability density function $p_{X^{PG'}}$ for $X^{PG'}$ that can be constructed based on $X^G$. This can be either a conditional density function, $p_{X^{PG'} | X^{PG}}$ -- or even $p_{X^{PG'} | X^{PG} \cup y^G}$ -- or a marginal density function, $p_{X^{PG'}}$. These multiple sampled values $x^{PG'}_{i,j}$ are then joined with the known values $x^{PG}_i$ for an instance after which $f^G$ is applied to all combinations of resampled and known values to obtain the corresponding labels for this instance. These are then aggregated into a soft label:

\small
\begin{equation}
\label{eq:soft_label}
y^{PG}_{i,c} = \sum\limits^{n}_{j=1} \frac{\mathds{1}_c[f^G(x^{PG}_i \cup [ x^{PG'}_{i,j} \sim p_{X^{PG'}}])]}{n},
\end{equation}
\normalsize

\noindent which returns the probability of class $c \in \mathcal{Y}$ for instance $i$, with $\mathds{1}_c$ the indicator function. As $n \rightarrow \infty$, an exact soft label for $y^{PG}_i$ is obtained given the selected probability density function. Feature hiding thus requires the actual values contained in $X^{PG}$, an assumed distribution over the variables in $X^{PG'}$ and the ground truth function $f^G$, so: $FH(f^G, x_i^{PG}, p_{X^{PG'}}) = y_i^{PG}$.

The values in $X^{PG'}$ could have been sampled from an infinite number of distributions and it is impossible to determine the exact original distribution. As we are constructing a new dataset, however, any distribution from which the values in $X^{PG'}$ and thus $X^G$ could possibly have been sampled is a valid choice. By selecting a specific distribution to resample from, the soft labels that reflect the ground truth given that specific distribution can be generated. $FH$ thus allows for the creation of a new $D^{PG}$ from $D^G$ by which the coupling between instances remains intact.

\subsubsection*{From Partial Ground Truth to Observed Soft Label}

The construction of $D^{OS}$ from $D^{PG}$ can be achieved through a number of non-invariant transformations. These transformations modify different components of $D^{PG}$ to produce the corresponding components of $D^{OS}$, thus creating experimental datasets tailored to various different classification tasks, including learning with label noise. The framework assumes that either the labels $y^{OS}$ are altered directly by the application of a noise model to $D^{PG}$, or indirectly by modifying the input variables, which ultimately leads to different labels being assigned by an annotator. The latter may be due to noise that was added into the observed input $X^O$ or $X^{O'}$ causing the data available to the annotator $X^A$ to differ from $X^{PG} \cup X^{PG'}$, or from the labelling function of the annotator $f^A$ being noisy. The pathways for introducing noise are listed below:

\begin{itemize}
\item $y^{PG}$ to $y^{OS}$: When $y^{OS}$ is not established based on the variables in $X^O$ and/or $X^{O'}$, but is measured directly, noise can be introduced directly to $y^{PG}$. This facilitates the NCAR and NAR noise models.
\item $X^{PG}$ and/or $X^{PG'}$ and possibly $y^{PG}$ to $y^{OS}$: If $y^{OS}$ is determined based on $X^{PG}$ and/or $X^{PG'}$ and $y^{PG}$, noise can be generated according to the NNAR model.
\item $X^{PG}$ to $X^{O}$ to $X^A$: By applying a non-invariant transformation, such as adding Gaussian noise to (some of) the variables in $X^{PG}$, $X^O$ can be altered directly. If $y^{OS}$ is decided upon by an annotator through $f^A$, using information $X^A$ which is based on $X^O$, the noise added to $X^{O}$ will be reflected in the annotated labels $y^{OS}$.
\item $X^{PG'}$ to $X^{O'}$ to $X^A$: If $y^{OS}$ is decided upon by an annotator through $f^A$ based on $X^A$, and this expert has more relevant information $X^{O'}$ available to them than is contained in $X^O$ alone, noise can be added to $y^{OS}$ by using a non-invariant transformation between $X^{PG'}$ and $X^{O'}$. For example $X^{O'}$ may include textual descriptions that a physician can use when annotating a dataset for the presence of some disease, containing information that might not be readily available for use by a classification model and is therefore not contained in $X^O$.
\item $X^A$ to $y^{OS}$: If, as in the previous two transformations, $y^{OS}$ is obtained trough $f^A$ based on $X^A$, the labels may be transformed by adjusting the annotation function $f^A$.
\end{itemize}

\noindent The specific transformation used to add label noise is decided upon by the user. The transformation should ensure that the resulting dataset is suited toward the classification task for which a method is to be examined. The advantage of generating the observed (noisy) data of interest by following these transformations down the chain, is that it guarantees the availability of the (Partial) Ground Truth labels for the coupled instances to be evaluated on. For research focusing on soft labels rather than label noise, the identity transformations may also be applied to $D^{PG}$, such that $X^O$ = $X^{PG}$ and $y^{OS}$ = $y^{PG}$ and a soft labelled observed dataset $D^{OS}$ is obtained.

\subsubsection*{From Observed Soft Label to Observed Hard Label}

Given $X^O$ and $y^{OS}$, the transformation to $D^{OH}$ is straightforward. A decision function $f_{dec}$ needs to be defined, which converts the soft labels $y^{OS}$ into hard labels $y^{OH}$. Examples of such functions include sampling the soft label distribution or simply selecting the class with the highest probability, though many more decision functions are possible. It is important to note that this function may be stochastic, allowing for random tie breaks in case of equal probabilities, as there exists no truth requirements for this functional relationship. 

\subsection{Back up}
\label{sec:up_the_chain}
  
Transformations down the chain, i.e. $D^G \rightarrow D^{PG} \rightarrow D^{OS} \rightarrow D^{OH}$, can be used to generate arbitrarily many observed datasets from a single ground truth dataset by adding various types of noise. However, obtaining a ground truth set is a prerequisite for this process. To generate a dataset $D^G$ using a real-world observed dataset, transformations up the chain have to be used. To ensure the instances remain coupled between datasets, such transformations have to respect the constraints that follow from Definition~\ref{def:G},\ref{def:PG},\ref{def:OS} and~\ref{def:OH}, summarized in Table~\ref{tab:dataset_restrictions}. Constructing the $D^G$ that corresponds to an existing set down the chain presents several difficulties:

\begin{itemize}
\item Transforming from $D^{PG}$ to $D^G$: A decision function $f_{dec}$ must be applied to $y^{PG}$ to transform the soft labels into hard labels. Unless $X^{PG'}$ is completely irrelevant or empty, making  $X^{PG}$ effectively equivalent to $X^G$, $X^{PG'}$ contains information that is required by $f^G$. This information results in different labels for $f_{dec}(f^G(X^{PG} \cup X^{PG'}))$ compared to $f_{dec}(f^G(X^{PG}))$ for some values in the domain $\mathcal{X}$ of $f^G$. Therefore, the missing information in $X^{PG'}$ causes this transformation to decouple the instances.
\item Transforming from $D^{OS}$ or $D^{OH}$ to $D^G$ or $D^{PG}$: As is required for the ground truth datasets, the true function $f^G$ for the task would need to be either known or learned. The latter is generally impossible, as observed data is finite and theoretically infinite functions describe it perfectly, while no further information is available to aid in the selecting the true governing function. Furthermore, duplicate instances in $X^O$ can have different hard or soft labels, whereas $y^G$ is constrained to contain deterministic hard labels.
\end{itemize}

\noindent The exception occurs when the sets along the chain are identical, requiring the observed sets to be noiseless and have a known ground truth function. These conditions rarely are rarely met in real-world datasets.

It is possible, however, to use SYNLABEL to construct a new, different ground truth dataset based on an observed dataset, such that its requirements are fulfilled. This will cause the instances in the new dataset to become decoupled from those in the original dataset. While this is an issue when the task is to find the best model for a specific dataset, it is of no concern when the aim is to create a suitable dataset for validation of a method under specific label noise conditions.

To construct a new $D^{PG}$, Definition~\ref{def:PG} requires that both soft labels and the true relationship between $X$ and $y$ are known. Since the latter generally cannot be discovered given $X^O$ and $y^{OS}$ or $y^{OH}$, we propose an alternative approach: first a function $f^O$ is learned based on $X^O$ and $y^{OS}$ or more commonly $y^{OH}$, assumed to produce probability estimates for each class (soft labels). This function is then set to be $f^{G}$, $X^{PG}$ to be $X^O$ and $y^{PG}$ to be $f^{G}(X^{PG})$. This effectively disregards the original labels and generates new soft labels by applying the selected ground truth function to the input data.

Furthermore, by imposing a deterministic decision function $f_{dec} $ on $f^{G}$, a new $D^G$ can be constructed from $X^{PG}$ by setting $y^G = f_{dec}(f^{G}(X^{PG}))$. Here, $f_{dec}$ has to be deterministic, so if for a binary problem $f_{dec}(f^{G}(X^{PG}))$ returns a probability of 0.5 for both classes, the decision may not be randomly taken. The truth function $f^G$ is altered as well: $f^G_{new} = f_{dec}(f^{G})$. This transformation again produces a new dataset consisting of different, decoupled instances. If the initially learned function $f^O$ already made deterministic classifications, the intermediate generation of $D^{PG}$ can effectively be skipped and $D^G$ can be generated directly from $X^O$ and $f^O$ instead. From the newly constructed $D^G$, any of the transformations described in Section~\ref{sec:down_the_chain} can be applied to construct any number of new datasets for experimentation.

The SYNLABEL framework, together with all of the described transformations, has been implemented and made available publicly on GitHub to encourage its use: \url{https://github.com/sjoerd-de-vries/SYNLABEL}.

\section{Application of the Framework}\label{sec:application}

In this section, we demonstrate the practical application of SYNLABEL for developing and evaluating noise-robust algorithms, highlighting its differences with existing methods for label noise experimentation. We refrain from comparing different algorithms on generated datasets -- one of the main uses of SYNLABEL -- as such comparisons would primarily reflect the performance of the algorithms rather than that of the framework. Throughout this study, we use The Keel Vehicle Silhouette set~\cite{triguero2017keel}, which consists of 18 features and 4 classes.

\subsubsection*{Constructing a Ground Truth}

To rigorously evaluate the performance of an algorithm, it is essential to have datasets for which the noise has been quantified. Some curated datasets are available, for which this has been done by experts. Not all noisy instances may have been identified, however, and they contain a fixed type of noise that might not align with our specific interests.
 
Alternatively, we could simulate clean datasets from scratch, e.g. by sampling data points from different normal distributions or constructing concentric circles, and then introduce the exact noise of interest. However, the resulting datasets typically do not capture the complexity of real-world data.    

Using SYNLABEL and the transformations up the chain defined in Section~\ref{sec:up_the_chain}, we can construct a $D^G$ based on observed, noisy data $D^O$. Typically, real datasets contain hard labels and as such we take $D^O = D^{OH} = \lbrace X^O, y^{OH} \rbrace$. A new dataset can then be constructed through the following steps: we train a machine learning model on $D^{OH}$, which acts as the function $f^O$. We set $X^G$ to be $X^O$. If $f^O$ produces soft labels we apply a deterministic decision function $f_{dec}$ to it and set the new ground truth function: $f^G = f_{dec}(f^O)$. Finally, we assign $y^G = f^G(X^O)$, to construct a  Ground Truth dataset inspired by real-world data.

The properties of this dataset depend on the model that is used as $f^G$. If a simple linear model is used, the resulting dataset will likely not contain the complexities expected to be present in a real-world system. On the other hand, if an overfitted neural network is used, the relationships may be overly complicated. As model expressivity is varied, baselines of corresponding complexity are constructed that approximate real-world problem difficulty to different extents.

\subsubsection*{Partial Ground Truth via Feature Hiding}

After generating $D^G$ as a baseline dataset for evaluation, we generate an additional set $D^{PG}$ with soft labels. Such a dataset allows for alternative, more direct ways of injecting and quantifying label noise compared to a set with hard labels, as discussed in Section~\ref{sec:quantification}.

$D^{PG}$ can be constructed by applying Feature Hiding, as specified by Equation~\ref{eq:soft_label}. First, we define which features to hide, i.e. add to $X^{PG'}$, and thereby which features remain in $X^{PG}$. Next, we specify the method for constructing the distribution from which the data in $X^{PG'}$ is resampled. Several methods are implemented in SYNLABEL and these can easily be extended to include custom methods. Finally, we specify the number of samples drawn and apply the transformation to obtain soft labels $y^{PG}$.

\begin{figure}[!t]
    \centering
    \includegraphics[width=0.47\textwidth]{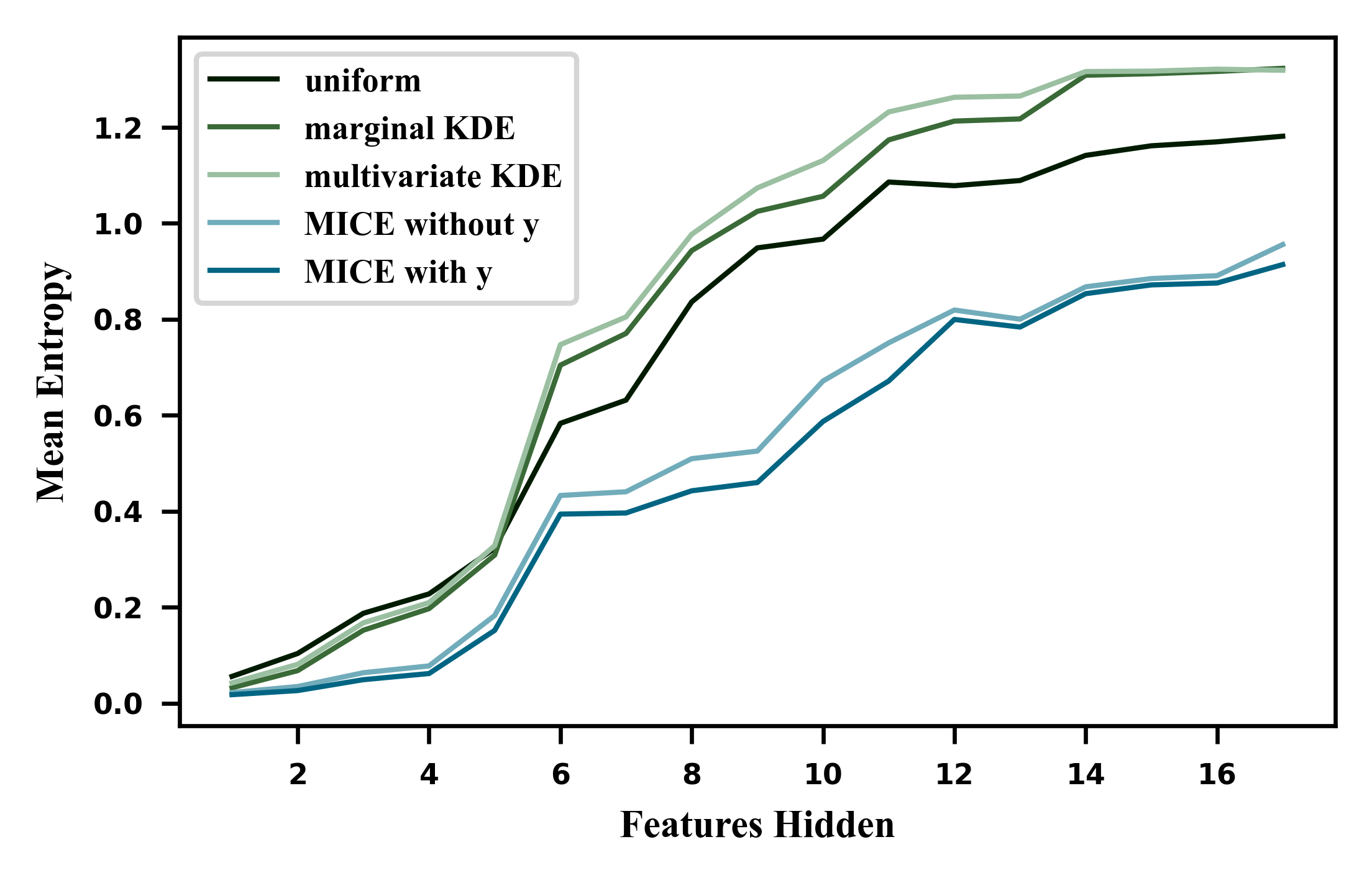}
    \caption{The level of label noise generated by Feature Hiding as measured by the mean entropy of the resulting soft labels for different probability density estimation methods and different numbers of features hidden. Average over 50 runs, with 100 values resampled for each feature. KDE: Kernel Density Estimation. MICE: Multivariate Imputation by Chained Equations.}
    \label{fig:feature_hiding}
\end{figure}

Figure~\ref{fig:feature_hiding} demonstrates how resampling from five prior distributions for $X^{PG'}$ constructed via different methods results in different levels of uncertainty in the obtained posterior distributions via Equation~\ref{eq:soft_label} (measured by the Shannon entropy) for different numbers of features hidden. As expected, sampling from a conditional probability density function $p_{X^{PG'} | X^{PG}}$ or even $p_{X^{PG'} | X^{PG} \cup y^G}$, in this case by using MICE~\cite{volker_vink_synthetic_mice_2021}, produces soft labels with lower entropy than sampling from a marginal probability density function $p_{X^{PG'}}$ does, as observed for the uniform and KDE methods.

\subsubsection*{Introducing Noise}

Having constructed both a baseline dataset with hard labels $D^G$ and soft labels $D^{PG}$, we proceed to introduce the label noise of interest to facilitate the comparative analysis of various methods on the resulting datasets. This noise can be added using any of the transformations detailed in Section~\ref{sec:down_the_chain} to generate $D^{OS}$. It is important to note that such noise injection can also be applied to an inherently noisy real-world dataset. However, the added noise would blend with any pre-existing, unspecified uncertainty, which makes it impossible to study the effects of the added noise in isolation, as demonstrated in Section~\ref{sec:quantification}. 

To obtain a noisy dataset with hard labels $D^{OH}$, a decision function $f_{dec}$ can be utilized to transform $y^{OS}$ into $y^{OH}$. This decision function could for example be to sample in proportion to the label distribution or to select the class with the highest probability.

\subsubsection*{Soft Label Learning and Annotators}

In addition to experimenting with label noise, SYNLABEL can be used to generate datasets for learning from soft labels or from labels assigned by one or more annotators.

An arbitrary number of soft labelled datasets $D^{PG}$, and by extension $D^{OS}$, can be constructed for any $D^G$ by varying which features are hidden (contained in $X^{PG'}$) and the distribution used for resampling in the Feature Hiding procedure. These datasets are valuable for specifically testing soft label learning algorithms, but they also can be used to address related problems. For instance, learning from confidence scores can be emulated by directly interpreting the soft labels $y^{PG}$ as a confidence distribution by setting them equal to $y^{OS}$, or additional noise can be introduced to model different levels of calibration for the confidences. Similarly, $y^{OS}$ can represent the aggregation of labels provided by members of the crowd to simulate a crowdsourcing scenario.

More elaborate scenarios for annotator labelling can be constructed by explicit use of the annotator function $f^A$. By constructing $y^{OS}$ based on $X^A$, it is possible to emulate scenarios in which the annotator has access to other features than the classifier via $X^{O'}$, a common scenario when expert annotators are involved. Furthermore, a consensus meeting or labelling by annotators with different levels of expertise can be emulated by adjusting $f^A$ accordingly.

\section{Uncertainty by Feature Hiding}\label{sec:fh_uncertainty}

The explicit definition of the ground truth function $f^G$ when using the SYNLABEL framework allows for the application of Feature Hiding ($FH$) to introduce uncertainty into the hard labels of $D^G$, resulting in a Partial Ground Truth set $D^{PG}$. Additionally, soft labels can be generated through alternative methods, such as randomly perturbing the one-hot encoded hard labels. In this section we demonstrate that Feature Hiding can be used to produce a different type of uncertainty with unique characteristics compared to transforming the labels with uniform or random probabilities, which corresponds to introducing noise using the NCAR or NAR models.

The repeated sampling of new values for the hidden features in the $FH$ method simulates a scenario in which not all information needed to classify instances perfectly for the task at hand is available. This is a common scenario for non-trivial classification problems, and the uncertainty it causes is known as epistemic uncertainty~\cite{hullermeier2021aleatoric}. In contrast, introducing uncertainty by assuming instances are altered by a predefined noise matrix falls into the category of aleatoric uncertainty.

To demonstrate that the uncertainty introduced by these different methods is indeed of a distinct nature, we conducted an experiment using the Keel Vehicle Silhouette dataset. Following the steps described in Section~\ref{sec:application}, we first set this dataset to be equal to $D^{OH} = \lbrace X^O, y^{OH} \rbrace$. Then, a Random Forest ($RF$) classifier was trained on this dataset and the Ground Truth dataset $D^{G}$ was defined by setting $X^G = X^{OH}$, $f^G = RF$ and $y^G = RF(y^{OH})$. Through each method -- $FH$, NCAR and NAR -- uncertainty was then introduced to generate a soft labelled set $D^{OS}$. For $FH$, this was accomplished by constructing a distinct Partial Ground Truth set $D^{PG}$ by hiding a number of features, which were added to $X^{PG'}$, while the remaining features were added to $X^{PG}$. The $FH$ transformation directly introduced the desired uncertainty into $y^{PG}$. By then applying the identity transformation ($X^{O'} = X^{PG'}, X^{O} = X^{PG}$ and $y^{OS}=y^{PG}$) we obtained $D^{OS}$. For NCAR and NAR, we first set $D^{PG} = D^G$ via the identity transformation, so $X^{PG'} = \emptyset, X^{PG} = X^{G}$ and $y^{PG}$ was equal to the one-hot encoded labels $y^G$. We then applied a noise matrix that matched the noise model to $y^{PG}$ to obtain $y^{OS}$, with $X^{O'} = X^{PG'} = \emptyset$ and $X^{O} = X^{PG}$. For each of the three methods, we generated both a soft labelled set $D^{OS}$ with low and high uncertainty in $y^{OS}$ and sampled the labels of the set with high uncertainty to obtain a hard labelled set $D^{OH}$.

To ensure a fair comparison between the different methods for introducing uncertainty, we kept the level of uncertainty in the resulting sets constant across these methods. We quantified the introduced uncertainty by measuring the Total Variation Distance, which for two discrete probability distributions (soft labels) $P$ and $Q$ is defined as:

\begin{equation}
TVD(P,Q) = \frac{1}{2} ||P - Q||_1.
\end{equation} 

To measure the uncertainty over the soft labels of a dataset, we used the mean $TVD$ or $\bar{TVD}$. A base level of uncertainty was established by setting the number of hidden features for $FH$ to 8 for low uncertainty and to 13 for high uncertainty. The features were ordered from lowest to highest feature importance as determined by the $RF$ model and the specified number of features was hidden following this order. The distribution used for resampling the hidden feature values was estimated from $X^{PG'}$ by multivariate Kernel Density Estimation. The $\bar{TVD}(y^G, y^{OS})$, with $y^G$ one-hot encoded, was measured to be 0.17 for the low level and 0.45 for the high level. The matrices used for the NCAR and NAR transformations were adjusted to match these uncertainty levels.

\begin{figure*}[p]
    \centering
    \includegraphics[width=\textwidth]{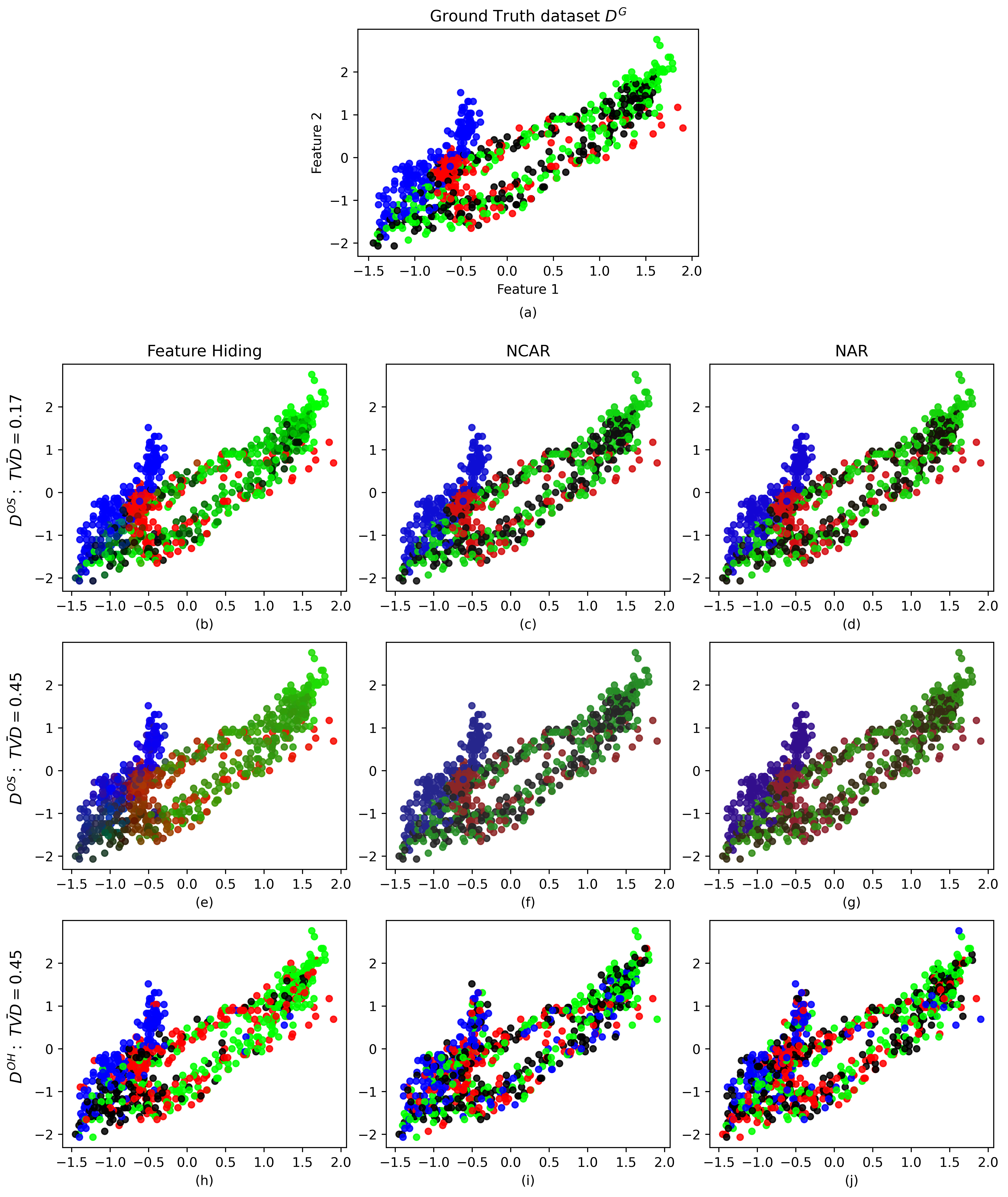}
    \caption{Different methods for introducing uncertainty. (a) The Ground Truth dataset $D^G$ generated from Keel Vehicle, using a Random Forest Classifier. In the remainder of the image: The left column (b,e,h) shows the result of applying Feature Hiding to the ground truth dataset. The middle column (c,f,i) shows the result of applying a uniform noise matrix to $D^G$. The right column (d,g,j) shows the result of applying a random noise matrix to $D^G$. The top row (b,c,d) shows the results for a level of uncertainty of $\bar{TVD}(y^{OS}, y^{G}) = 0.17$, the middle row (e,f,g) for $\bar{TVD}(y^{OS}, y^{G}) = 0.45$, and the bottom row (h,i,j) the result of sampling a hard label from the soft labels from the middle row to obtain $D^{OH}$. }
    \label{fig:uncertainty}
\end{figure*}

The results of this experiment are shown in Figure~\ref{fig:uncertainty}. In the plots the outcome labels are displayed in red, blue, green or black for the hard labels. For soft labels, the values for each of the individual classes in a label was translated to a value in the RGB color space, with the black class equal to the absence of color. Along the horizontal and vertical axes the values of two of the remaining five features, that were not hidden for the high uncertainty scenario, are shown. At the top (Fig.~\ref{fig:uncertainty}a) the Ground Truth dataset with corresponding hard labels is shown. 

At the lower uncertainty level (Fig.~\ref{fig:uncertainty}b,c,d), we observe distinct patterns across different methods. Specifically, for Feature Hiding (Fig.~\ref{fig:uncertainty}b), the black class becomes less pronounced in the upper and right regions of the feature space. Where hard labels previously were black, we now frequently observe a mixture of the green and black labels. In contrast, For NCAR (Fig.~\ref{fig:uncertainty}c) and NAR (Fig.~\ref{fig:uncertainty}d) the color distribution appears consistent with the Ground Truth labels. All colors have become slightly more blurred, however.

At the higher uncertainty level (Fig.~\ref{fig:uncertainty}e,f,g), the patterns that we observed for the lower level have become more clear. For Feature Hiding (Fig.~\ref{fig:uncertainty}e), the black class has become most present in the bottom left region of the feature space, while the red and green labels have become slightly more uncertain, they remain clearly distinguishable. The blue class has become a little more uncertain as well. For NCAR (Fig.~\ref{fig:uncertainty}f) and NAR (Fig.~\ref{fig:uncertainty}g) the color distribution from $D^G$ remains clearly visible, but colors are more blurred then before. To quantify how the the soft labels have changed, we measured the entropy of the resulting label distributions. We found that for $FH$, the average entropy was 0.90. For the uniform and random transformations they were 1.18 and 1.15 respectively. As they share the same level of $\bar{TVD}$, and thus how much the labels have transformed from $D^G$ to $D^{OS}$, we can conclude that for $FH$ the changes moved the labels more from one class to another, whereas the NCAR and NAR methods caused all labels to move more towards a uniform probability distribution, for which all labels are equally likely and the entropy is maximal. 

The sampled labels $y^{OH}$ (Fig.~\ref{fig:uncertainty}h,i,j) allow for a direct comparison with $y^G$ from $D^G$. The analyses for the high level of uncertainty hold, as we observe the class labels to have moved for $FH$ (Fig.~\ref{fig:uncertainty}h), whereas they have mainly become more randomly distributed for NCAR (Fig.~\ref{fig:uncertainty}i) and NAR (Fig.~\ref{fig:uncertainty}j). Interesting to note, however, is that for NCAR the blue class seems more present throughout the feature space than for NAR. This can be explained as for NCAR all class transition probabilities are equal, whereas for NAR they are randomly generated, which caused there to be an overall lower transition probability to the blue class.

In summary, Feature Hiding offers a distinct method of generating uncertainty compared to other methods such as introducing the uncertainty randomly via the NCAR or NAR models. The uncertainty $FH$ introduces can be explained as due to incomplete information (epistemic uncertainty) and it causes the class distributions to change accordingly, as opposed to them just becoming more uniformly distributed across the feature space (aleatoric uncertainty).

\section{Quantifying Label Noise}\label{sec:quantification}

An inherent advantage of the availability of soft labels is that uncertainty can be quantified by measures such as the Shannon entropy, or when two (discrete) probability distributions are to be compared, the Total Variation Distance. The latter enables us to introduce label noise directly into the soft labels of $D^{PG}$ to generate $D^{OS}$ and precisely quantify the resulting noise for each individual label by evaluating $TVD(y_i^{PG}, y_i^{OS})$. This cannot be achieved for hard labels, where noise is binary -- it either changes the labels or it does not -- and can only be estimated by repeated application of the noise injection method. 

Quantifying noise also necessitates a baseline dataset for comparison. When using SYNLABEL to evaluate classifier performance under selected noise conditions, either the probabilistic output of the classifier can be compared directly to $y^{PG}$, or its hard labels to $y^G$ or $y^{PG}$. Furthermore, the effects of the introduced noise can be analysed both independently and in combination with any pre-existing uncertainty due to partly unobserved data. In contrast, when noise is introduced into an already noisy dataset, the two sources of uncertainty become indistinguishable.

\begin{figure*}[!t]
\centering
\begin{subfigure}{0.5\textwidth}
  \centering
  \includegraphics[width=1\textwidth]{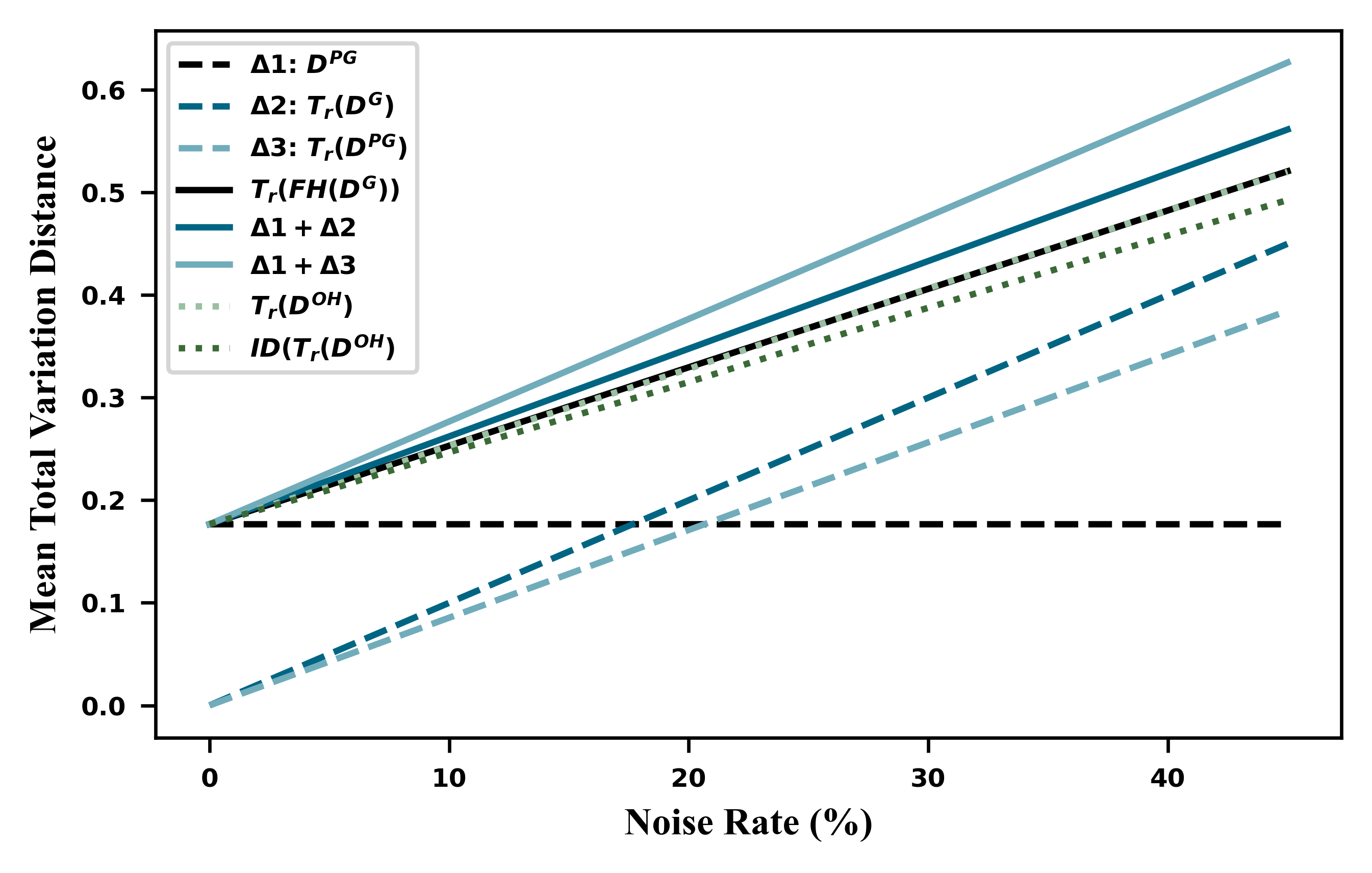}
\end{subfigure}%
\begin{subfigure}{0.5\textwidth}
  \centering
  \includegraphics[width=1\textwidth]{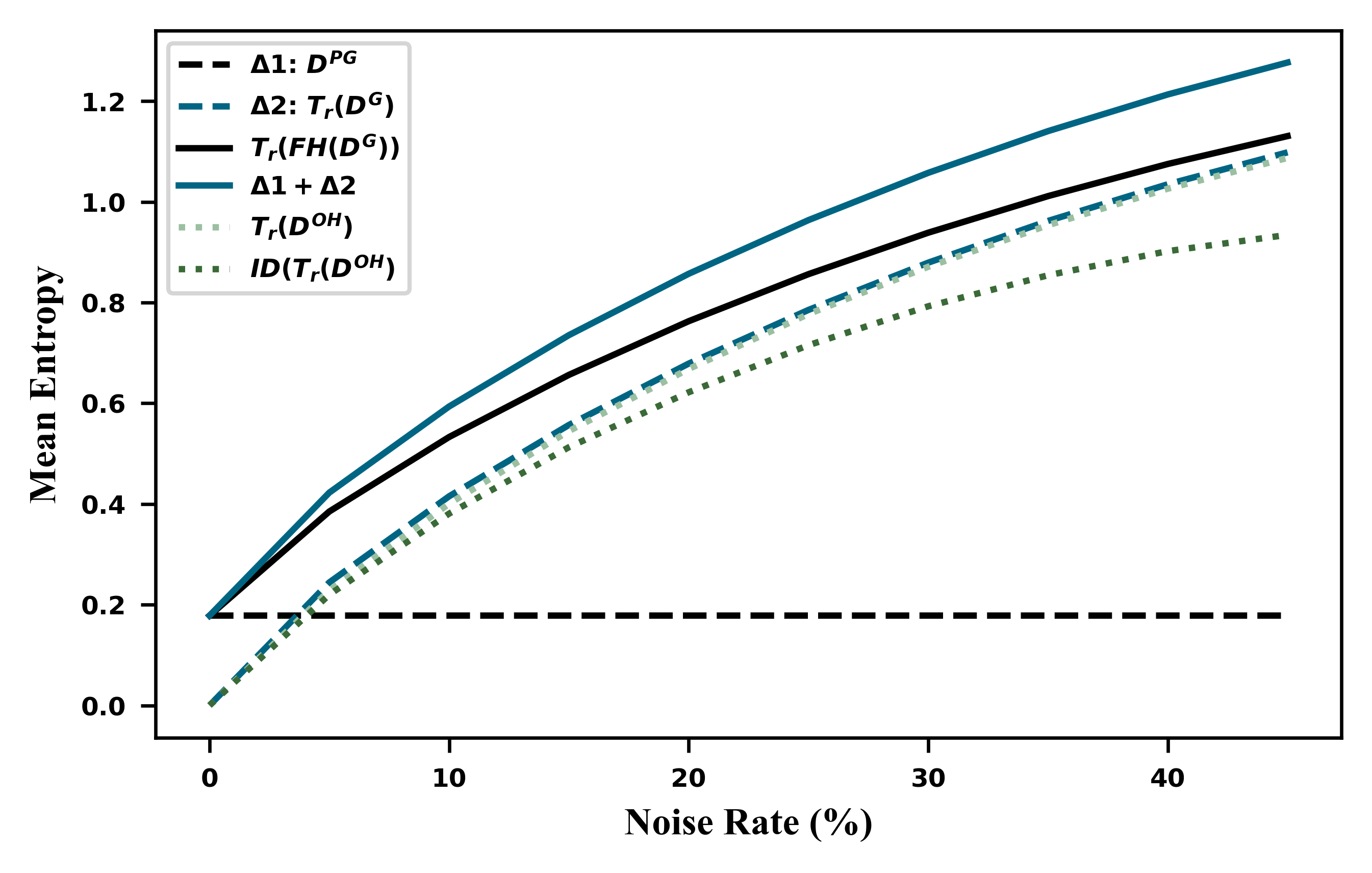}
\end{subfigure}
\caption{Different noise measures for varying noise rates. Left: the mean $TVD$. Feature Hiding was done by sampling from a marginal distribution constructed via Kernel Density Estimation (KDE). Uniform noise (NCAR) was added by applying noise matrix $T_r$. Right: the mean entropy. Feature Hiding was done by sampling from a conditional distribution using MICE. Random class-conditional noise (NAR) was introduced by a randomly generated $T_r$, with equal probabilities on the main diagonal. $T_r$: transition matrix. ID: instance-dependent (NNAR). $FH$: Feature Hiding. $\Delta_1$: noise introduced by $FH$. $\Delta_2$: noise introduced by applying $T_r$ to $D^G$. $\Delta_3$: noise introduced by applying $T_r$ to $D^{PG}$.}
\label{fig:noise_injection}
\end{figure*}

To demonstrate both the importance of a clean baseline, i.e. a baseline without noise or with uncertainty that is fully understood, and how label distributions enable better quantification of noise and direct noise injection, we conducted two experiments with the Keel Vehicle Silhouette dataset. The results are presented in Figure~\ref{fig:noise_injection}. In these experiments, we first constructed $D^G$ based on this observed dataset $D^{OH}$, by training a Random Forest classifier on the original data, setting $f^G$ equal to this classifier, $X^G$ to $X^O$ and $y^G$ to $f^G(X^O)$. We then added uncertainty via Feature Hiding ($FH$) and/or additional noise to this baseline. The strength of this noise varies along the horizontal axis, indicated by the probability that a label is transformed from its original label into another class by the corresponding noise matrix. The level of uncertainty resulting from this procedure is shown on the vertical axis.

On the left side of the figure, the mean Total Variation Distance ($\bar{TVD}$) is depicted, either with respect to the one-hot encoded labels in $D^G$ or, in case of $\Delta_3$, with respect to $D^{PG}$. This metric measures the change within the individual label probabilities of one soft label compared to another soft label, averaged over two related sets of soft labels. For generating $D^{PG}$ through $FH$, a distribution for $X^{PG'}$ was estimated using marginal Kernel Density Estimation.

We observe a difference between the level of uncertainty when applying NCAR through a uniform noise matrix $T_r$ to a $D^{PG}$ that has been constructed via $FH$ from $D^G$, denoted as $T_r(FH(D^G))$, and the separately added uncertainty of $\Delta_1 = D^{PG} = FH(D^G)$ and $\Delta_3 = T_r(D^{PG})$, denoted as $\Delta_1 + \Delta_3$. The same applies when this noise is added to $D^G$ directly, denoted as $\Delta_2 = T_r(D^G)$, and then added to the uncertainty introduced by $FH(D^G)$, denoted as $\Delta_1 + \Delta_2$. This difference in uncertainty, resulting from the same noise matrix being applied to different datasets, illustrates how label noise applied to a set with unknown baseline uncertainty cannot be retrospectively isolated and properly quantified.

$T_r(D^{OH})$ is the result of sampling hard labels from $D^{OS}=D^{PG}$ (100 times) and then applying the uniform flipping probability via $T_r$ to each sample (100 times as well), while for $T_r(FH(D^G)$ the noise matrix is simply applied to $y^{PG}$ directly. As desired, $T_r(D^{OH}) = T_r(FH(D^G))$, demonstrating that the direct injection of noise into the label distribution makes repeated sampling from $D^{PG}$, followed by repeated application of random noise functions to individual instances, redundant. This approach saves 10.000 repeated actions and ensures the measured noise level is exact instead of an estimation. In addition, some types of noise are more naturally applied directly to a label distribution. 

Finally, $ID(T_r(D^{OH}))$, where ID stands for instance-dependent (NNAR), is included to illustrate that similar results are obtained for a more complex type of noise. The uniform $T_r$ from before is used, but it is applied twice as often to the instances with the largest ratio of distance to their nearest neighbour of the same label to distance to a neighbour of another label, as described in Garcia et al.~\cite{garcia2019new}.

On the right side of the figure, the results of a similar experiment in which the uncertainty is measured by the mean entropy of the label distributions are shown. Here, instead of uniform noise, $T_r$ is a class-conditional noise (NAR) matrix. MICE was used, without taking $y^G$ into account, for resampling values in the $FH$ method to add further variation compared to the previous experiment. Since entropy is not a measure between distributions, $\Delta_3$ cannot be measured with respect to $y^{PG}$ and is therefore omitted, as the entropy is already measured by $T_r(FH(D^G))$. The instance-dependent noise is shown for reference. The same patterns are observed here as for $\bar{TVD}$: adding the uncertainty introduced by the separate steps of Feature Hiding ($\Delta_1 = D^{PG} = FH(D^G)$) and noise injection ($\Delta_2 = T_r(D^G)$), i.e. $\Delta_1 + \Delta_2$, clearly results in a higher level of uncertainty than adding the noise after $FH$, i.e. $T_r(FH(D^G))$, does. This confirms that noise added to pre-existing, unknown uncertainty cannot be studied in isolation. 

Once again, the uncertainty generated by sampling hard labels from $D^{OS} = D^{PG}$ and then randomly changing the hard labels $D^{OH}$ through many repeated actions is equivalent to applying the transition matrix to the soft labels directly. Furthermore, the latter gives an exact result instead of an approximation and saves many repeated actions.

\section{Conclusion}\label{sec:conclusion}

This work introduces SYNLABEL, a framework specifically designed for constructing synthetic data for label noise experiments. SYNLABEL features methods to create a ground truth dataset inspired by real-world data. This is achieved by learning a classification function from the real-world data, establishing it as the ground truth function for the new dataset, and then applying it to the original input data to generate new hard labels. Furthermore, SYNLABEL can transform this dataset into a partial ground truth set with soft labels through a procedure called Feature Hiding. This method involves hiding certain input features contained in the domain of the selected function, resampling values from learned or pre-specified distributions for these hidden features, evaluating the ground truth function on the resulting data and aggregating the obtained labels. Feature Hiding simulates uncertainty due to missing information, a scenario often encountered in classification problems. The introduced uncertainty is shown to fundamentally differ from that produced by conventional approaches. Together these two types of ground truth datasets provide a clean baseline into which any noise of interest may be introduced, enabling the precise quantification of such noise as well as the thorough evaluation of its effect on method performance. 

Conducting experiments using datasets generated by SYNLABEL offers significant advantages over the three types of datasets typically used in label noise research. Firstly, the generated data are more complex than data that are simulated from scratch. Secondly, the framework ensures the availability of clean baseline data for evaluation, which are often unobtainable for real-world data. Finally, the noise being studied can be precisely specified, unlike for curated datasets, that contain fixed noise and are costly to construct. Furthermore, SYNLABEL is applicable to soft label learning and related classification tasks such as learning from crowds, expert data and data with confidences scores. In summary, SYNLABEL proves to be a versatile framework that enables researchers to construct synthetic data that can be used for experiments across a variety of fields.

\bibliography{references}

\end{document}